\documentclass[letterpaper, 10 pt, conference]{ieeeconf} 
\IEEEoverridecommandlockouts

\usepackage[letterpaper, left=0.75in, right=0.75in, bottom=0.8in, top=0.75in]{geometry}

\usepackage{cite}
\usepackage{lipsum}
\usepackage{booktabs}
\usepackage{amsmath,amssymb,amsfonts}
\usepackage{algorithmic}
\usepackage{graphicx}
\usepackage{textcomp}
\usepackage{xcolor}
\usepackage{multirow}
\usepackage{mwe}
\usepackage{hyperref}
\usepackage{bm}

\DeclareMathOperator*{\argmax}{argmax}
\DeclareMathOperator*{\argmin}{argmin}

\begin{document}

\title{SmartBelt: A Wearable Microphone Array for\\ Sound Source Localization with Haptic Feedback}

\author{Simon Michaud, Benjamin Moffett, Ana Tapia Rousiouk, Victoria Duda, Fran\c{c}ois Grondin
\thanks{This work was supported by FRQNT -- Fonds recherche Qu\'ebec Nature et Technologie. The authors would like to thank Jean-Gabriel Mercier and Gabriel Dor\'e for their assistance with the experiments.}
\thanks{S. Michaud, B. Moffett and F. Grondin are with the Department of Electrical Engineering and Computer Engineering, Interdisciplinary Institute for Technological Innovation (3IT), 3000 boul. de l'Universit\'e,
Universit\'e de Sherbrooke, Sherbrooke, Qu\'ebec (Canada) J1K 0A5, A.T. Rousiouk and V. Duda are with the School of Speech-Language Therapy and Audiology, Université de Montréal, 7077 Av du Parc, Montréal, Qu\'ebec (Canada) H3N 1X7 \texttt{\{simon.michaud}, \texttt{benjamin.moffett}, \texttt{francois.grondin2}\}\texttt{@usherbrooke.ca}, \texttt{\{ana.tapia.rousiouk}, \texttt{victoria.duda\}@umontreal.ca}}}

\maketitle

\begin{abstract}
This paper introduces SmartBelt, a wearable microphone array on a belt that performs sound source localization and returns the direction of arrival with respect to the user waist.
One of the haptic motors on the belt then vibrates in the corresponding direction to provide useful feedback to the user.
We also introduce a simple calibration step to adapt the belt to different waist sizes.
Experiments are performed to confirm the accuracy of this wearable sound source localization system, and results show a Mean Average Error (MAE) of 2.90$^{\circ}$, and a correct haptic motor selection with a rate of 92.3\%.
Results suggest the device can provide useful haptic feedback, and will be evaluated in a study with people having hearing impairments.
\end{abstract}

\section{Introduction}

Hearing assistive devices such as bone-anchored hearing devices and cochlear implants have a limited ability to locate and separate sounds \cite{verschuur2005auditory,denk2019limitations, nelissen2016three}.
This is an issue as localization is an important auditory ability for safely moving in space and detecting sounds in environments such as schools, restaurants, busy streets and workplaces. Hearing in noise is facilitated by the localization of sounds in order to turn the head towards the speaker (the source of the sound) and utilize additional speech cues such as lip-reading and facial expressions to extract a message from high background noise.
Haptic devices have been shown to significantly improve the localization of sounds \cite{fletcher2020electro}.
Studies have also shown that auditory localization can be improved by training \cite{byrne1998optimizing}.
People lacking access to visual information are particularly reliant on their auditory localization capacities to move  autonomously and safely \cite{guth2010perceiving,lagrow1994orientation, lawson2010improving,kolarik2016auditory}.

Robot audition offers interesting solutions to perform sound source localization using microphone arrays in dynamic environments.
The existing approaches are well-suited for assisting a human with sound localization tasks as: 1) microphone arrays are available in arbitrary geometries as robots come in many different shapes; and 2) they can run on portable low-cost hardware powered by a battery.
There are three main types of sound source localization methods: 1) steered-power beamformers; 2) subspace decomposition methods; 3) machine learning-based approaches.

Steered-Response Power with Phase Transform (SRP-PHAT) consists in steering a beamformer in different potential direction of arrivals (DoAs) around the microphone array \cite{dibiase2001robust}.
It is also possible to break down the problem for each pair of microphones, using the Generalized Cross-Correlation with Phase Transform (GCC-PHAT) \cite{brandstein1997robust}.
The computational load however increases with the number of microphones and the number of potential DoAs \cite{grondin2013manyears}.
To reduce the complexity, the search can first be done on a coarse grid, and then finished with a finer grid \cite{grondin2019lightweight}.
Another approach is the Singular Value Decomposition with Phase Transform (SVD-PHAT), which relies on Singular Value Decomposition of the SRP-PHAT projection matrix to project the observation on a vector with the principal components \cite{grondin2019svd,grondin2019fast}.
A k-d tree is then used to speed up the search for the most likely direction of arrival of sound.
These approaches require the \emph{a priori} geometry of the microphone array to be known, which can be an issue if the microphones are installed on a deformable surface.

Subspace decomposition consists of methods such as MUltiple SIgnal Classification (MUSIC) \cite{schmidt1986multiple} and Estimation of Signal Parameters via Rotational Invariance Technique (ESPRIT) \cite{roy1989esprit}.
While MUSIC was formerly used with narrowband signals, it was adapted for wideband signals such as speech with the Standard Eigenvalue Decomposition MUSIC (SEVD-MUSIC) algorithm \cite{ishi2009evaluation}.
SEVD-MUSIC assumes that speech is more powerful than noise for each frequency bin, which is not always the case.
The Generalized Eigenvalue Decomposition with MUSIC (GEVD-MUSIC) overcomes this limitation by performing decomposition with respect to the noise spatial correlation matrix \cite{nakamura2009intelligent,nakamura2011intelligent,nakadai2012robot}.
GEVD-MUSIC however suffers from the non-orthogonality of the bases spanning the noise subspace, which in turn impacts the accuracy of localization.
Generalized Singular Value Decomposition with MUSIC (GSVD-MUSIC) is proposed to enforce bases orthogonality and improve localization accuracy \cite{nakamura2012real}.
MUSIC-based methods however rely on eigenvalue or singular value decomposition in real-time, which involves a significant amount of computations, and usually require expensive hardware.

Machine learning approaches have recently gained in popularity for a wide range of audio processing tasks, including multi-channel sound source localization.
They usually involve training a neural network for a specific microphone array geometry, based on simulated and/or recorded audio data.
For instance, it is shown that localization with a uniform linear array can be performed by training a convolutional neural network (CNN) using white noise signals \cite{chakrabarty2017broadband, chakrabarty2019multi}.
Convolutive and Recursive Neural Networks (CRNNs) can also be used to estimate DoAs for a specific class of sounds \cite{adavanne2018sound,adavanne2018direction}.
Although these deep learning methods show accurate DoA estimation, they have two major drawbacks: 1) they need a significant amount of training data for a specific microphone array geometry, which makes quick calibration difficult; and 2) they require expensive hardware to perform inference as they have numerous weight parameters. 

In this paper, we propose a new device, called SmartBelt, that can be installed on a user waist.
This belt performs sound source localization with multiple microphones using GCC-PHAT with binary time-frequency masks, and provides haptic feedback using motors to inform the user regarding the direction of arrival of sound.
There are three main contributions in this paper: 1) we introduce and describe the first belt that performs audio localization and provides haptic feedback; 2) we propose a simple yet effective calibration for the belt to adapt to different waist sizes; and 3) we demonstrate that this belt can estimate DoA accurately and provide useful haptic feedback.
This paper is organized as follows.
Section \ref{sec:smartbelt} introduces the hardware and the proposed algorithm.
Section \ref{sec:experiments} demonstrates how the belt performs in real conditions.
Finally, section \ref{sec:conclusion} concludes with final remarks and suggests future work.

\section{SmartBelt}
\label{sec:smartbelt}

The proposed SmartBelt device consists of custom made hardware and uses algorithms based on GCC-PHAT to estimate the DoA of the sound source of interest.

\subsection{Hardware}

\begin{figure}[!ht]
    \centering
    \includegraphics[width=0.9\linewidth]{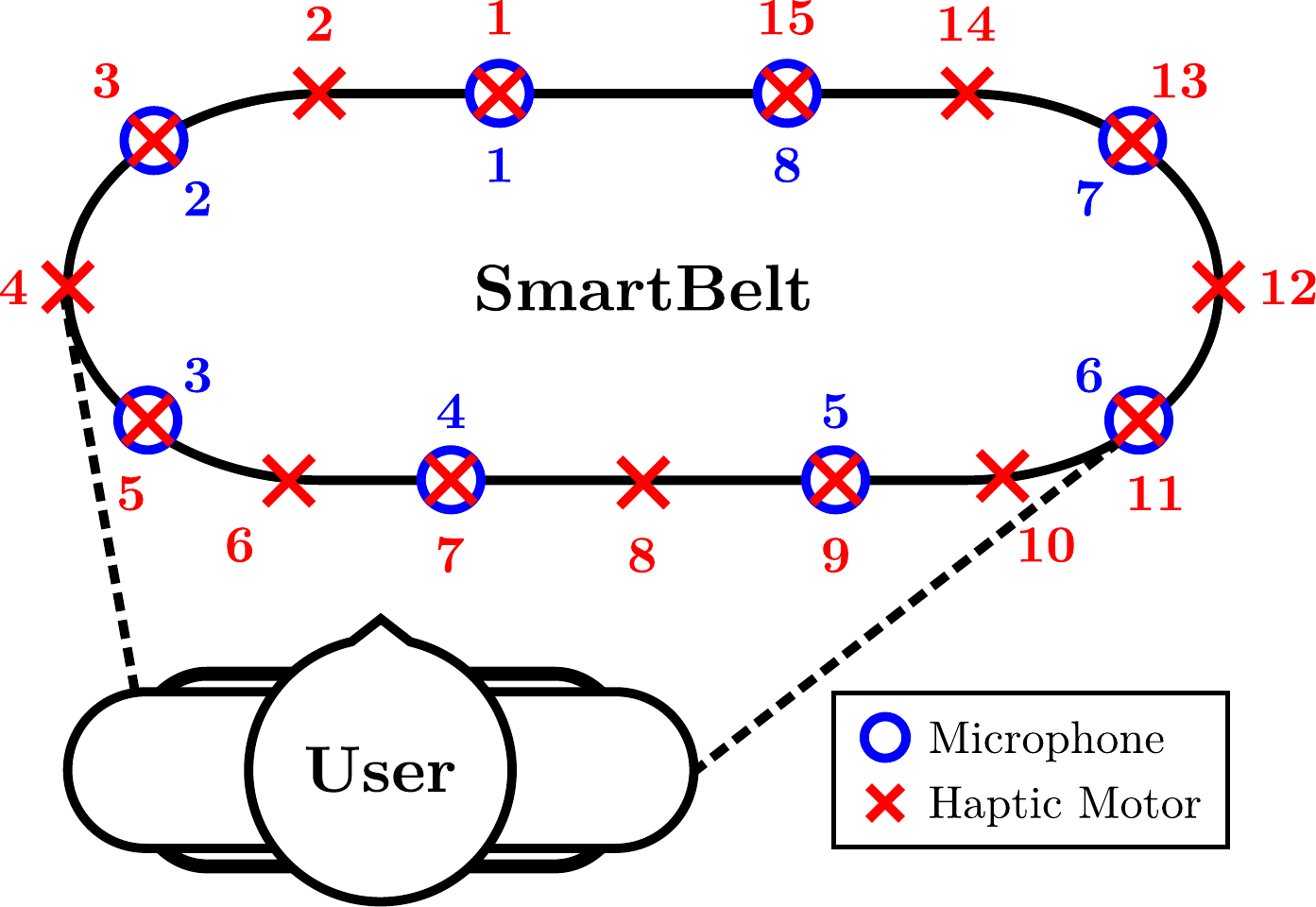}
    \caption{SmartBelt Layout}
    \label{fig:belt}
\end{figure}

The proposed hardware consists of 8 microphones fixed on the belt perimeter and 15 haptic motors.
The haptic motors are fixed to the internal side of the belt, and a microphone is also installed every two motors on the external side of the belt on 3-D printed supports, as shown in Figure \ref{fig:belt}.
The geometry of the microphone array is unknown \emph{apriori}, but it is assumed that the motors are spaced evenly on the belt.

\begin{figure}
    \centering
    \includegraphics[width=\linewidth]{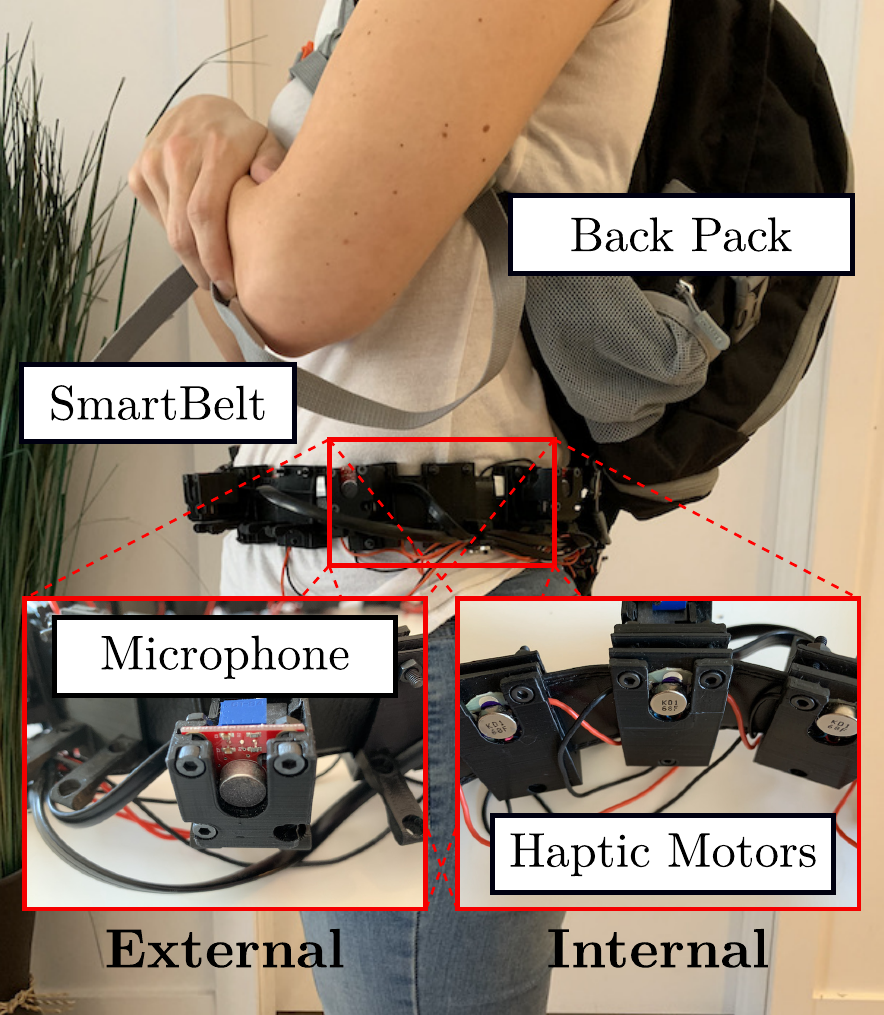}
    \caption{The belt is installed around the user's waist. The microphones are connected to a sound card in the bag, and the haptic motors are driven by mosfet boards also installed in the back pack.
    Cables leave the bag and run along the belt perimeter to reach each motor and microphone individually.}
    \label{fig:hardware}
\end{figure}

\begin{figure}
    \centering
    \includegraphics[width=\linewidth]{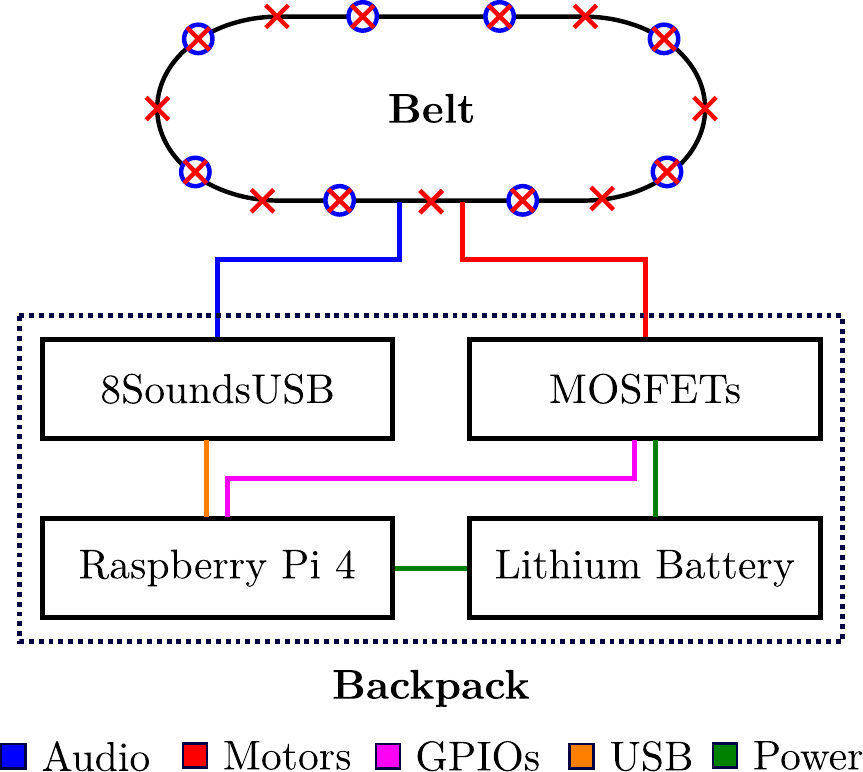}
    \caption{Diagram with the connections between the different hardware components of the SmartBelt}
    \label{fig:schematic}
\end{figure}

The microphones are plugged in a 8SoundsUSB sound card \cite{abran2014usb}, connected to a Raspberry Pi 4 (RP4) board \cite{mcmanus2021raspberry} via USB.
Each haptic motor is connected to a MOSFET module, which is interfaced with a General Purpose Input/Output (GPIO) pin on the RP4.
A portable power bank with USB ports powers the RP4 and the MOSFETs.
The sound card, MOSFETs, RP4 and power bank are installed in a backpack carried by the user.
Cables exit the backpack and connect to the belt from behind.
Figure \ref{fig:hardware} shows the hardware for SmartBelt, and Figure \ref{fig:schematic} illustrates the connection schematic.

\subsection{Algorithms}

\subsubsection{TDoA Estimation}
\label{subsubsec:tdoaest}

The proposed system captures the audio signals from the microphones and compute a Short-Time Fourier Transform (STFT) for each canal $c \in \mathcal{C} = \{1, 2, \dots, 8\}$ over windows of $N \in \mathbb{N}$ samples for a total of $T \in \mathbb{N}$ frames, denoted as $X_c(t,f) \in \mathbb{C}$, where $t \in \{1, 2, \dots, T\}$ stands for the frame index and $f \in \{0, 1, \dots, N/2\}$ stands for the frequency bin index.
The cross-correlation $R(t,f) \in \mathbb{C}$ is computed in the frequency domain for a each pair of microphones $(u,v) \in \mathcal{Q}$ (where $\mathcal{Q}=\{(x,y) \in \mathcal{C}^2: x<y\}$) as follows:
\begin{equation}
    R_{u,v}(t,f) = X_u(t,f)X_v(t,f)^*,
\end{equation}
where $(\dots)^*$ stands for the complex conjugate.
The Time Difference of Arrival (TDoA) estimation for each pair of microphones $(u,v)$ used here relies on the Generalized Cross-Correlation with Phase Transform (GCC-PHAT) approach, which makes use of the inverse Fast Fourier Transform (iFFT) to estimate the correlation in the time domain as follows:
\begin{equation}
    r_{u,v}(t,\tau) = \sum_{f=0}^{N-1}{M(t,f)\frac{R_{u,v}(t,f)}{|R_{u,v}(t,f)|}\exp\left(-j2\pi f\tau/N \right)},
\end{equation}
where $\tau \in \mathbb{Z}$ corresponds to the time delay in samples, $j = \sqrt{-1}$ and $|\dots|$ stands for the magnitude of the complex number.
A binary time-frequency mask $M(t,f) \in \{0,1\}$ is generated to capture only regions different from background noise (estimated during silence periods), as previously investigated in \cite{grondin2015time,grondin2016noise}.
The TDoA $\tau_{u,v}(t)$ at each frame $t$ is then obtained by finding the peak in the correlation signal:
\begin{equation}
    \tau_{u,v}(t) = \argmax_{\tau}{\{r_{u,v}(t,\tau)\}}.
\end{equation}

All the TDoAs are accumulated in a buffer of $T$ frames, and the final TDoA corresponds to the mode:
\begin{equation}
\tau_{u,v} = \mathrm{Mod}(\tau_{u,v}(t)).    
\end{equation}

As the exact geometry of the microphone array is unknown, a calibration step is required to map the TDoAs of all pairs of microphones to each potential DoA, and assign a specific DoA for each haptic motor position.

\subsubsection{Calibration}

Each person has a unique morphology (waist circumference, width, etc.), which can also change over time due to health related issues (e.g. pregnancy, sedentary life style, aging).
For this reason, a fast and quick calibration procedure with minimal hardware requirement is desirable.
The goal is to associate each haptic motor to a DoA azimuth angle (in degrees).

The user first positions himself at an approximate angle, and then a loudspeaker a few meters away plays a white noise signal ant the same height and for 3 seconds.
The belt records the audio signals, estimates the TDoA for each pair of microphones, and stores them in memory.
The user then orients himself in another direction, and the same procedure is repeated.
The eight calibration directions are set to $0^{\circ}$, $45^{\circ}$, $90^{\circ}$, $135^{\circ}$, $180^{\circ}$, $225^{\circ}$, $270^{\circ}$ and $315^{\circ}$.
The system then estimates the TDoAs for each of the 360 angles (with a resolution of 1 degree) by performing linear interpolation with the eight calibration directions and associated TDoAs.
A lookup table is then generated with each DoA and the corresponding $28$ TDoAs:

\begin{equation}
    \left[
    \begin{array}{c}
    0 \\
    1 \\
    \vdots \\
    359 \\
    \end{array}
    \right]
    \rightarrow
    \left[
    \begin{array}{ccccc}
    \tau^0_{1,2} & \tau^0_{1,3} & \tau^0_{1,4} & \dots & \tau^0_{7,8} \\
    \tau^1_{1,2} & \tau^1_{1,3} & \tau^1_{1,4} & \dots & \tau^1_{7,8} \\
    \vdots \\
    \tau^{359}_{1,2} & \tau^{359}_{1,3} & \tau^{359}_{1,4} & \dots & \tau^{359}_{7,8} \\
    \end{array}
    \right].
\end{equation}

\begin{figure}[!ht]
    \centering
    \includegraphics[width=\linewidth]{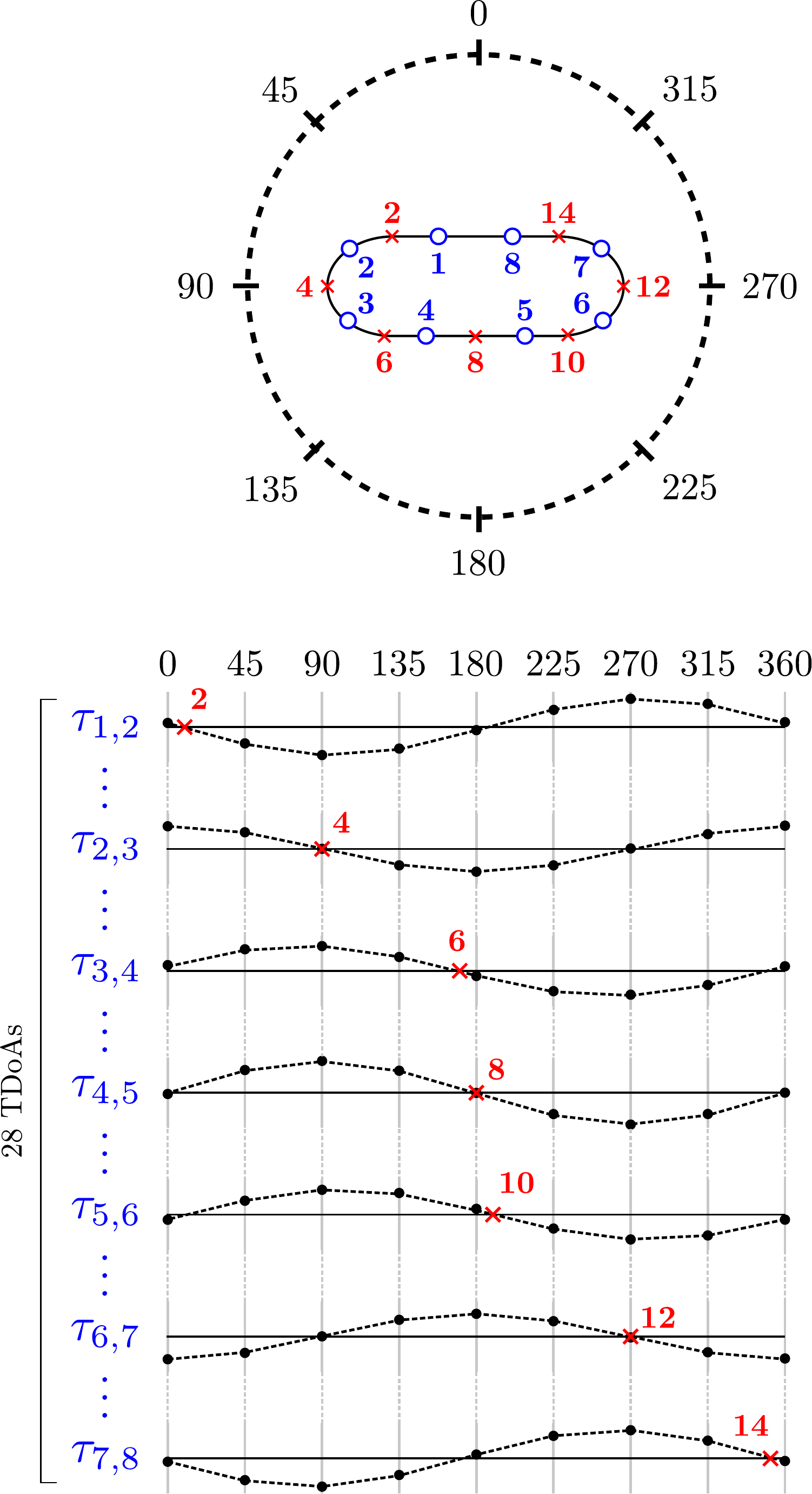}
    \caption{A sound source generates white noise at eight different angles around the belt ($0^{\circ}$, $45^{\circ}$, $\dots$, $315^{\circ}$).
    The TDoAs are estimated for each calibration angle, and are connected using linear interpolation. The DoA angles aligned with the haptic motors with even indices correspond to the zero-crossing for pairs of microphones $(1,2)$, $(2,3)$, $(3,4)$, $(4,5)$, $(5,6)$, $(6,7)$ and $(7,8)$. Only the TDoAs that correspond these 8 pairs out of 28 are shown for clarity. In this example, the DoA associated to haptic motor $6$ is $\theta_6 = 170^{\circ}$}.
    \label{fig:tdoas_haptic}
\end{figure}

The haptic motors $\{2$, $4$, $6$, $8$, $10$, $12$, $14\}$ are positioned midway between the pair of microphones $\{(1,2)$, $(2,3)$, $(3,4)$, $(4,5)$, $(5,6)$, $(6,7)$, $(7,8)\}$.
This implies that when the sound source faces directly one of these motors, the TDoA estimated at the corresponding pair of microphones corresponds to $0$ (both microphones are equidistant from the motor).
Using this property, it is possible to associate a DoA angle (between $0^{\circ}$ and $360^{\circ}$) to each haptic motor with an even index.
We denote these angles as $\theta_2$, $\theta_4$, $\theta_6$, $\theta_8$, $\theta_{10}$, $\theta_{12}$ and $\theta_{14}$.
Figure \ref{fig:tdoas_haptic} shows how the DoA of the haptic motors can be estimated.
It is interesting to note that the calibration angles can be more or less accurate (i.e. they can differ from the exact values of $0^{\circ}$, $45^{\circ}$, $\dots$, $315^{\circ}$).
In fact, the interpolated angles can be slightly off and generate a reference scale that is imperfect, but the predicted DoA at test time and the reference haptic motor DoAs are both on this same scale, which compensates for the difference.

The haptic motors $3$, $5$, $7$, $9$, $11$ and $13$ are equidistant to neighbor motors which DoAs are known.
Using linear interpolation, the DoAs can be computed as follows:
\begin{equation}
\begin{array}{ccc}
    \displaystyle\theta_3 = \frac{\theta_2 + \theta_4}{2},&
    \displaystyle\theta_5 = \frac{\theta_4 + \theta_6}{2},&
    \displaystyle\theta_7 = \frac{\theta_6 + \theta_8}{2},\\
    \\
    \displaystyle\theta_9 = \frac{\theta_8 + \theta_{10}}{2},&
    \displaystyle\theta_{11} = \frac{\theta_{10} + \theta_{12}}{2},&
    \displaystyle\theta_{13} = \frac{\theta_{12} + \theta_{14}}{2}.\\
\end{array}
\end{equation}

Finally, the DoAs of the two last haptic motors $1$ and $15$ are estimated as follows:
\begin{equation}
    \theta_1 = \frac{3\theta_2 - \theta_3}{2},\ \theta_{15} = \frac{3\theta_{14} - \theta_{13}}{2}.\ 
\end{equation}

\subsubsection{Localization}

Once the belt is properly calibrated, it can be used to localize the DoA of a sound stimulus that would normally draw the attention of a person with normal hearing function (e.g. phone ringing, car honking).
Once the belt localizes the sound of interest, the haptic motor that matches this direction vibrates to provide feedback to the user.

The TDoAs are computed the same way as in Section \ref{subsubsec:tdoaest}.
The score between the computed TDoAs ($\hat\tau_{u,v}$) and the TDoAs in the lookup table ($\tau^\phi_{u,v}$) for each potential DoA $\phi$ generated during calibration is obtained using a Squared Exponential Kernel (to make it more robust to potential outliers):
\begin{equation}
    f(\phi) = \sum_{u=1}^{8}\sum_{v=u+1}^{8}\sigma^2 \exp\left(-\frac{( \hat\tau_{u,v}-\tau^\phi_{u,v} )^2}{2l^2}\right).
\end{equation}
The predicted DoA $\phi^*$ then corresponds to the potential DoA with the highest score:
\begin{equation}
    \phi^* = \argmax_{\phi} \left\{ f(\phi) \right\},
\end{equation}
and the belt activates the haptic motor with the closest DoA $\theta_i$ to the predicted DoA $\phi^*$:
\begin{equation}
    \theta_i^* = \argmin_{i} \left\{ |\theta_i - \phi^*| \right\}.
\end{equation}

\section{Experiments}
\label{sec:experiments}

The belt is tested in controlled real-life conditions to demonstrate the versatility of the proposed system amongst different users.
The frame ($N$) and hop ($\Delta N$) sizes are chosen to ensure a $23$ msec analysis window with an overlap of $50\%$.
The Squared Exponential Kernel parameters $l$ and $\sigma$ are chosen empirically to provide good localization accuracy.
Table \ref{tab:doa} shows the parameters used for the experiments.

\begin{table}[!ht]
    \renewcommand{\arraystretch}{1.4}
    \centering
    \caption{Parameters for DoA estimation}
    \begin{tabular}{|c|c|}
        \hline
        Parameter & Value \\
        \hline
        $N$ & 1024 \\
        $\Delta N$ & 512 \\
        $l$ & 0.707 \\
        $\sigma$ & 1 \\
        \hline
    \end{tabular}
    \label{tab:doa}
\end{table}

\begin{figure}[!ht]
    \centering
    \includegraphics[width=0.9\linewidth]{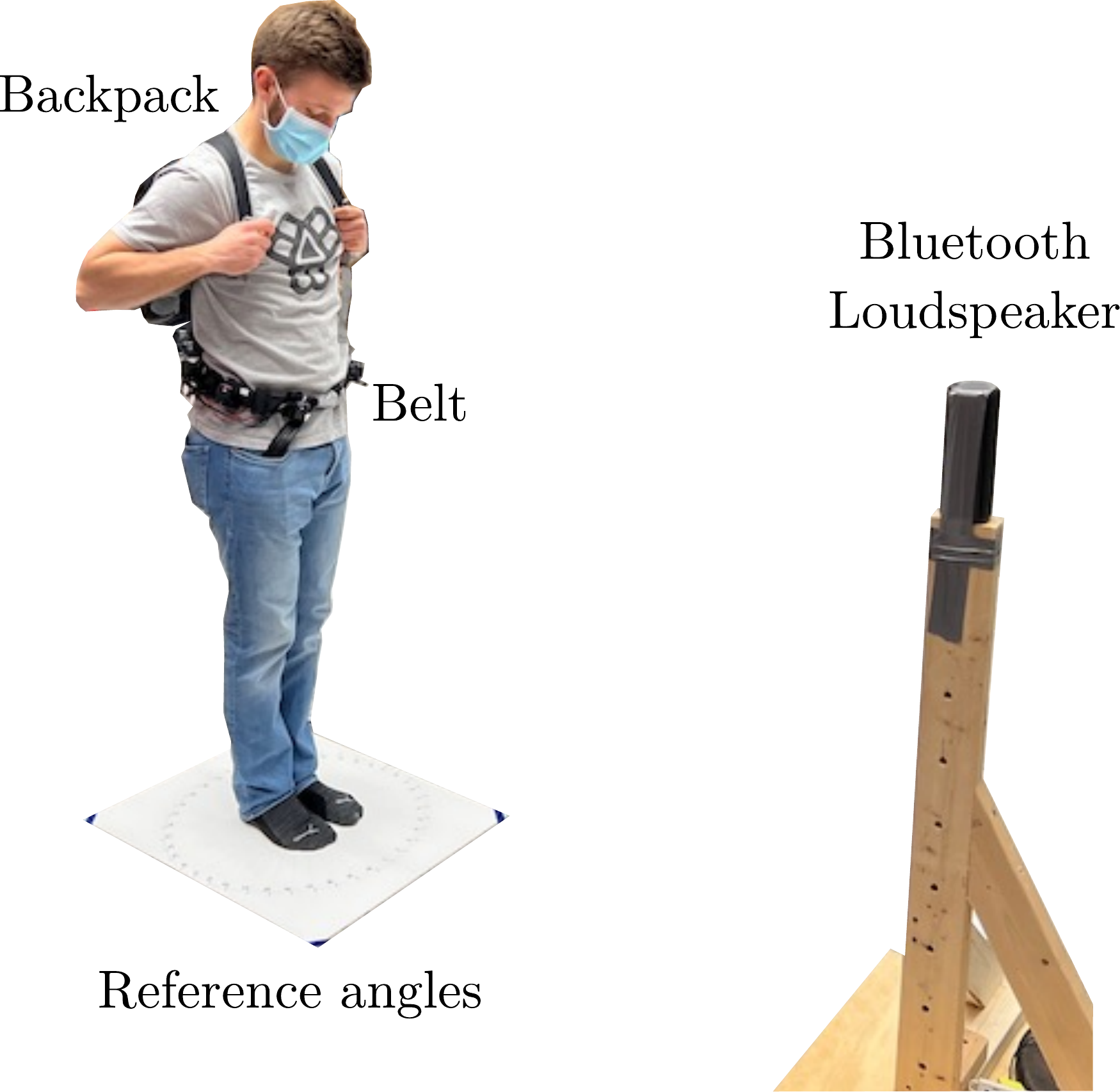}
    \caption{Setup for the experiment with a participant wearing the belt and the loudspeaker playing sounds.}
    \label{fig:exp_setup}
\end{figure}

A bluetooth speaker is used as the sound source, and a general purpose cardboard with 40 azimuth angles ($0^{\circ}$, $9^{\circ}$, $18^{\circ}$, $\dots$, $342^{\circ}$, $351^{\circ}$) serves as a reference for the orientation of the user. The speaker is approximately 2 m from the user and at the same height. 
For testing,  multiple sounds of interests are used: 1) truck horn; 2) car driving by; 3) car horn; 4) car braking; 5) phone ringing; 6) speed car accelerating.
Each sound has a duration of approximately 2 seconds.
The DoA estimation is performed using the full segment.
The belt is worn by two male participants with different waist sizes.
Calibration is performed by playing white noise at eight angles ($0^{\circ}$, $45^{\circ}$, $\dots$, $315^{\circ}$).
The test sounds are then played at each one of the 40 azimuth angles.
Figure \ref{fig:exp_setup} demonstrates the experimental setup used to evaluate the performance of the belt.

Using the calibration procedure described earlier, the DoAs associated to each haptic motor are obtained and shown in Table \ref{tab:motors}.
The results demonstrate that it is possible to easily recover the positions of the haptic motors using the proposed calibration method as they are moved on the belt to accommodate for the different waist sizes. 

\begin{table}[!ht]
    \renewcommand{\arraystretch}{1.4}
    \centering
    \caption{Corresponding DoA angles for haptic motors}
    \begin{tabular}{|c|cc|}
    \hline
        Motor & Participant A & Participant B \\
    \hline
        1 & 36$^{\circ}$ & 30$^{\circ}$ \\
        2 & 51$^{\circ}$ & 45$^{\circ}$\\
        3 & 80$^{\circ}$ & 75$^{\circ}$\\
        4 & 109$^{\circ}$ & 105$^{\circ}$\\
        5 & 134$^{\circ}$ & 127$^{\circ}$\\
        6 & 159$^{\circ}$ & 150$^{\circ}$\\
        7 & 169$^{\circ}$ & 168$^{\circ}$\\
        8 & 180$^{\circ}$ & 187$^{\circ}$\\
        9 & 195$^{\circ}$ & 202$^{\circ}$\\
        10 & 210$^{\circ}$ & 218$^{\circ}$\\
        11 & 235$^{\circ}$ & 239$^{\circ}$\\
        12 & 260$^{\circ}$ & 261$^{\circ}$\\
        13 & 282$^{\circ}$ & 279$^{\circ}$\\
        14 & 305$^{\circ}$ & 298$^{\circ}$\\
        15 & 316$^{\circ}$ & 307$^{\circ}$\\
    \hline
    \end{tabular}
    \label{tab:motors}
\end{table}

The test sounds are played at each position on the loudspeaker, and the predicted ($\phi$) and reference ($\gamma$) DoAs angles are compared.
The Mean Absolute Error (MAE) is then computed for each test sound and participant as follows:
\begin{equation}
    \mathrm{MAE} = \frac{1}{40}\sum_{k=1}^{40}{|\phi_k-\gamma_k|},
\end{equation}
where $\phi_k$ and $\gamma_k$ stand for the predicted and baseline angle at position $k$, where $k \in \{1, 2, \dots, 40\}$.

Table \ref{tab:mae} shows the MAE for each sound and participant.
The results confirm the accuracy of the proposed method to estimate the DoA.
On average, both MAEs are similar, with values of $2.69^{\circ}$ and $3.11^{\circ}$ for participants A and B, respectively, and an overall average of $2.90^{\circ}$.
This is similar to MAE in humans, estimated in some studies to 3 degrees for wideband sounds coming from a specific direction \cite{blauert1997spatial}.

\begin{table}[!ht]
    \renewcommand{\arraystretch}{1.4}
    \centering
    \caption{MAE for each sound and participant}
    \begin{tabular}{|c|cc|}
        \hline
        Sound & Participant A & Participant B \\
        \hline
        Truck horn & 4.55$^{\circ}$ & 3.25$^{\circ}$ \\
        Car driving by & 2.45$^{\circ}$ & 2.83$^{\circ}$ \\
        Car horn & 2.23$^{\circ}$ & 4.80$^{\circ}$ \\
        Car braking & 2.65$^{\circ}$ & 2.25$^{\circ}$ \\
        Phone ringing & 2.50$^{\circ}$ & 3.25$^{\circ}$ \\
        Speed car accelerating & 1.73$^{\circ}$ & 2.25$^{\circ}$ \\
        \textbf{Average} & \textbf{2.69$^{\circ}$} & \textbf{3.11$^{\circ}$} \\
        \hline
    \end{tabular}
    \label{tab:mae}
\end{table}

Based on the predicted DoAs, we choose the haptic motor that needs to provide feedback.
The chosen motor is compared to the one that should be activated given the theoretical DoA, and a ratio of good match is computed.
Table \ref{tab:percentage} shows these results.
In general, the belt provides a high fidelity haptic feedback with most sounds, with an average of $92.5\%$ and $92.1\%$ for participants A and B respectively, and an overall average of $92.3\%$.
Note that the lowest performances are observed with the truck horn sound, which is expected as for this sound segment, most of the power lies in the low frequencies, and GCC-PHAT performs better with wideband signals.

\begin{table}[!ht]
    \renewcommand{\arraystretch}{1.4}
    \centering
    \caption{Proportion of good match between activated haptic motors and haptic motors that should be activated}
    \begin{tabular}{|c|cc|}
    \hline
        Sound & Participant A & Participant B \\
    \hline
        Truck horn & 77.5\% & 82.5\% \\
        Car driving by & 100.0\% & 90.0\% \\
        Car horn & 95.0\% & 87.5\% \\
        Car braking & 92.5\% & 100\% \\
        Phone ringing & 95.0\% & 95.0\% \\
        Speed car accelerating & 95.0\% & 97.5\% \\
        \textbf{Average} & \textbf{92.5\%} & \textbf{92.1\%} \\        
    \hline
    \end{tabular}
    \label{tab:percentage}
\end{table}


\section{Conclusion}
\label{sec:conclusion}

This paper presents three contributions: 1) we introduce the first belt that can perform sound source localization at $360^{\circ}$ around a user; 2) we propose a simple calibration procedure to adjust the belt for participants with different waist sizes; and 3) we demonstrate that the belt provides accurate DoA estimation and haptic feedback.
So far, a simple time-frequency mask is applied to detect non-stationary sound source and ignore stationary noise.
In practice however, it would be important to trigger the haptic feedback only for sounds of interest.
To achieve this, joint sound source detection and localization could be applied to each pair of microphones \cite{grondin2019sound}.
Moreover, the performance of GCC-PHAT deteriorates with narrower bandwidth sounds, as observed in the experiments with the truck horn sound.
It would therefore be useful to combine the inter-level difference between microphones with the delay of propagation information.
Moreover, a new prototype could be built with MEMS microphones connected in daisy chain, which would reduce the amount of wiring, reduce the power consumption, and make the device more portable.

Given the challenges associated with the use of only amplification to transmit auditory localization information, one potential application of the SmartBelt would be to couple it to a hearing aid device or cochlear implant. In this way, the haptic motors could contribute additional vibrotactile stimulation complementing the auditory information transmitted via the hearing device. Currently there is little information in the literature regarding the benefits of multi-modal stimulation for the improvement of auditory capacities such as localization. Future work could include measuring the functional consequences of vibrotactile stimulation without amplification on a normal-hearing population. Later experiments could be carried out on clinical populations with hearing losses of various degrees coupled to different types of hearing devices. Combining multimodal stimuli has been shown to improve reaction time, perceptual precision and accuracy \cite{burr2009temporal,ernst2004merging,fujisaki2005temporal,camponogara2016hear,mclachlan2021towards} which is particularly important when navigating a physical environment such as a busy street. Future experiments will look at the degree to which threshold auditory localization perception can be improved by combining haptic and auditory stimulation.

\bibliography{bibliography}
\bibliographystyle{IEEEtran}

\end{document}